\documentclass[11pt,a4paper]{article}
\usepackage[hyperref]{emnlp2020}
\usepackage{times}
\usepackage{latexsym}

\usepackage{stmaryrd}
\usepackage{algorithm}
\usepackage{algorithmicx}
\usepackage{algpseudocode}
\usepackage{booktabs}
\usepackage{graphicx}
\usepackage{amssymb}
\usepackage{amsmath}
\usepackage{sidecap}
\usepackage{adjustbox}
\usepackage{wrapfig}
\usepackage{caption}
\usepackage{float}
\usepackage[smallerops]{newtxmath}

\usepackage{enumitem}
\DeclareMathAlphabet{\mathcal}{OMS}{cmsy}{m}{n}
\DeclareMathAlphabet{\mathbb}{U}{msb}{m}{n}

\newcommand{\wh}{\emph{wh}}
\newcommand{\role}[3][0]{\underset{\textsc{#3}}{\underline{\textrm{#2}}}^{#1}}
\newcommand{\best}[1]{\textbf{{#1}}}
\newcommand{\squad}{SQuAD}
\newcommand{\squadtwo}{SQuAD 2.0}

\usepackage{microtype}

\aclfinalcopy 

\title{Reading the Manual: Event Extraction as Definition Comprehension}

\author{
  Yunmo Chen$^{1}$ \quad Tongfei Chen$^{1}$ \quad Seth Ebner$^{1}$ \\ \bf Aaron Steven White$^{2}$ \quad Benjamin Van Durme$^{1}$ \\
  $^{1}$Johns Hopkins University \quad $^{2}$University of Rochester \\
  \texttt{\string{yunmo,tongfei,seth,vandurme\string}@jhu.edu} \\
  \texttt{aaron.white@rochester.edu}
}

\date{}

\begin{document}
\maketitle
\begin{abstract}
We ask whether text understanding has progressed to where we may extract event information through incremental refinement of \emph{bleached statements} derived from annotation manuals.  Such a capability would allow for the trivial construction and extension of an extraction framework by intended end-users through declarations such as, \emph{Some person was born in some location at some time}.  We introduce an example of a model that employs such statements, with experiments illustrating we can extract events under closed ontologies and generalize to unseen event types simply by reading new definitions.
\end{abstract}

\setlist[itemize]{leftmargin=*}  
\setlist[1]{itemsep=0pt}

\section{Introduction}

\begin{figure}
      \centering
      \includegraphics[width=\linewidth]{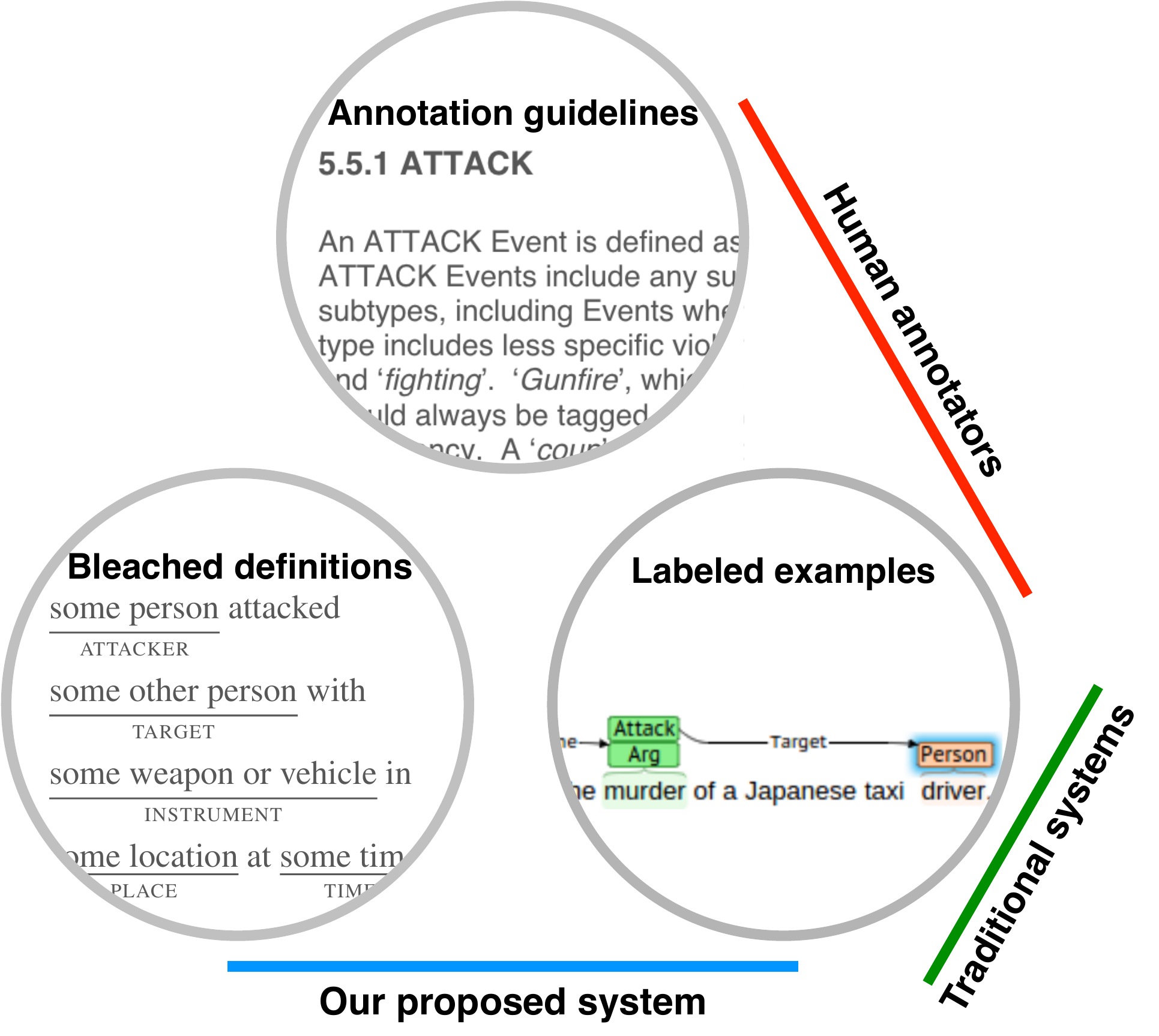}
      \caption{Comparison of data sources for human annotators, traditional information extraction systems, and our proposed approach. Human annotators use annotation guidelines and limited illustrative examples, traditional systems use large amounts of labeled examples, and our system uses bleached statements (derived from annotation guidelines) and labeled examples.}
      \label{fig:data-source} 
\end{figure}

This work is aimed at the disconnect between how human annotators and machines carry out information extraction: humans read annotation manuals consisting of guidelines and illustrative examples then label data, whereas machines label data (by making predictions) based purely on previously seen examples (Figure~\ref{fig:data-source}). 
%
%
We explore the feasibility of building a model that has access to information derived from annotation manuals.  Specifically we focus on the task of event extraction and convert annotation guidelines describing event types into natural language \emph{bleached statements}. An example bleached statement for the ACE 2005~\cite{ACE2005} \textsc{Life:Be-Born} event type is:
  \begin{equation}\resizebox{\columnwidth}{!}{$
  \role[]{some person}{person} ~{\textrm{was born in}}~ \role[]{some location}{place} ~{\textrm{at}}~ \role[]{some time}{time} \nonumber$}
  \end{equation}
  
 The bleached statement describes a general occurrence of an event of a given type. The event's arguments are initialized with bleached placeholders (e.g. \emph{some person}) to be replaced with extracted spans from the text, eventually resulting in, e.g.:
  \begin{flalign*}
    &\role[]{Barack Obama}{person} ~{\textrm{was born in}}~ \role[]{Hawaii}{place} ~{\rm at}~ \role[]{some time}{time} \nonumber
  \end{flalign*}

  We are motivated to consider these statements owing to the rapid progress in sentence-level representation learning and machine reading comprehension: can a contemporary encoder \emph{understand} event statements  well enough that their derived representation may be directly employed in extraction?
      
  \begin{figure*}[t]
    \centering
    \includegraphics[width=\linewidth]{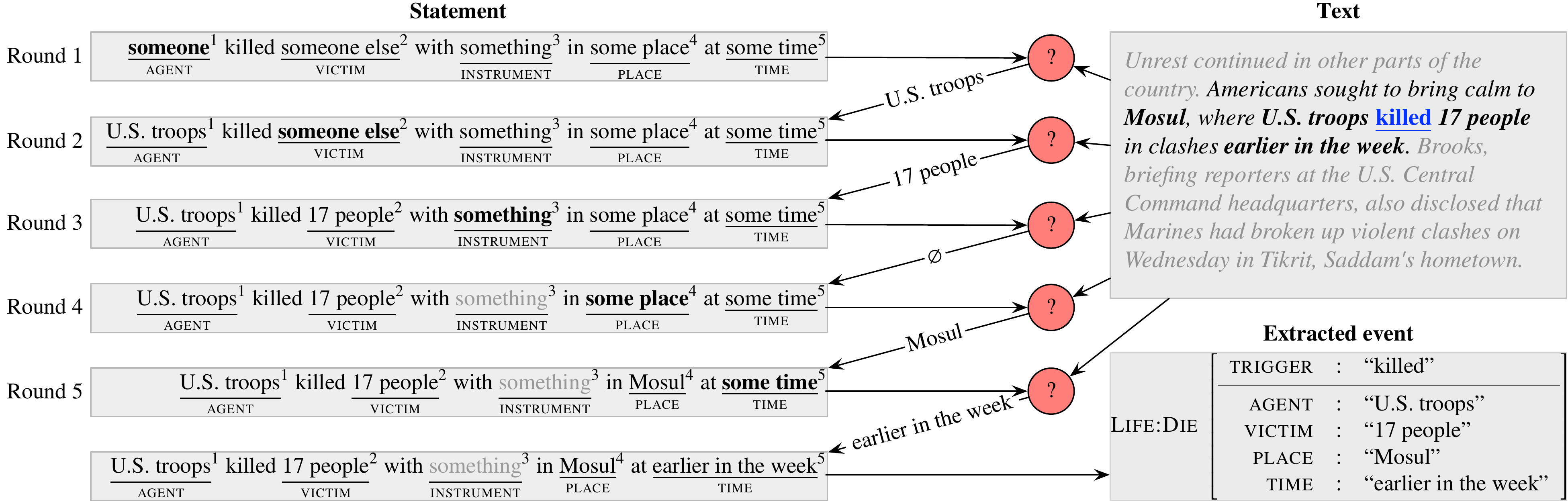}
    \vspace{-0.5cm}
    \caption{An example of our approach on a sentence from the ACE 2005 dataset for the \textsc{Life:Die} event. The bleached statement is incrementally populated with values from the text (in the order denoted by the superscripts), and not all event arguments are supported by the text. The grayed out text in the paragraph is given for context to the reader, but our model operates on single-sentence contexts.}
    \label{fig:statement-completion}
    \vspace{-0.2cm}
  \end{figure*}
  Bleached statements are straightforward to write and accommodate various levels of expressiveness in both the choice of arguments present and in the lexicalization of the trigger.\footnote{For example, for a \textsc{Conflict:Attack} event, we could use the simple trigger ``attacked'' or the more descriptive phrase ``violently caused physical harm or damage to.''} These features allow for easy adaptation as an ontology changes: simply introduce new or modified statements.  In the case where the amount of labeled examples is small or non-existent, 
  a bleached statement serves as a sort of canonical example provided ahead of further annotation.  We would like a solution where one simply declared the intended information, where any traditional labeled examples were used for disambiguation and fine-tuning, rather than being a critical component in building a model.
  
  
  As an example of such a solution, we propose a model which incrementally populates bleached statements by querying partially filled statements against text. This strategy is similar to the tasks of machine reading comprehension (MRC) and question answering (QA), in which an answer span is predicted in response to a question about a document.  Conceptually this may also be considered a form of incremental recognizing textual entailment (RTE) \cite{rte-1} where we iteratively refine a hypothesis that is supported by the document context.
  %
  %
  Experimental results demonstrate that zero- and few-shot event extraction are feasible with this approach.  While our intent here is exploratory, we manage to 
  achieve state-of-the-art performance on trigger identification and trigger classification on the ACE 2005 dataset. 
  %
  The contributions of this work are: {\bf (i)} A novel approach to event extraction that takes into account annotation guidelines through bleached statements; {\bf (ii)} A multiple-span selection model that demonstrates the feasibility of the approach for event extraction as well as for zero- and few-shot settings.

\section{Background}

  Event extraction is traditionally viewed as three subtasks: (1) event trigger detection, where triggers of events (words that most clearly express the occurrences of events) are detected; (2) entity mention detection, where all potential arguments (entity mentions) to events are detected; and (3) argument role prediction, where relations between detected arguments and trigger words are recognized with respect to each event type's defined set of roles.
  
  Much prior work adopts a pipelined approach to these 3 subtasks or focuses on a subset of the subtasks based on gold entity mention spans. These include feature-based approaches~\cite[\textit{inter alia}]{ji-grishman-2008-refining,liao-grishman-2010-using,McClosky11,HuangR12,li-etal-2013-joint} and neural approaches~\cite[\textit{inter alia}]{NguyenG15,ChenXLZ015,ChenLZLZ17,NguyenG18,ShaQCS18}. 
  
  Because pipelined approaches suffer from error propagation in which the error from earlier subtasks (e.g. entity mention detection) is inherited by later subtasks, joint modeling of the 3 subtasks has been attempted.
  \citet{yang-mitchell-2016-joint} attempts to jointly model the three components with hand-crafted features, but still need to detect entity mentions and event triggers separately. \citet{NguyenN19} jointly models the three tasks using neural networks with shared underlying representations. The models proposed in these two works are the baselines used in this paper.
  
  \citet{huang-etal-2018-zero} approach zero-shot event extraction by stipulating a graph structure for each event type and finding the event type graph structure whose learned representation most closely matches the learned representation of the parsed AMR \cite{BanarescuBCGGHK13} structure of a text. In contrast, our approach forgoes explicit graph-structured semantic representations such as AMR.  

  Researchers have introduced large question answering (QA) / machine reading comprehension (MRC) datasets in a cloze style~\cite{HermannKGEKSB15,OnishiWBGM16}, where a query sentence contains a placeholder and the model fills the blank. Our work can be viewed as an extension to such work, where multiple placeholders are extracted.
  
  \citet{li-etal-2019-entity} casts relation extraction as multi-turn QA with natural language questions, where in each turn one argument of the relation is found.. The method requires writing a  question for each entity type and each relation type.
  In~\cite{levy-etal-2017-zero}, sets of crowdsourced paraphrastic questions are written for each relation type in the ontology. In contrast, for each event type we use a single declarative bleached statement derived from the annotation guidelines.
  \citet{SoaresFLK19} proposes a model for relation extraction by filling in two blanks given a contextual relation statement.

  These three methods focus on binary relation extraction, and do not readily generalize to $n$-ary events or relations. Our approach naturally supports variable arity events and relations. 
  
\section{Problem Formulation}
  A \emph{bleached statement} consists of: the statement tokens $ S = (s_1, s_2, \cdots, s_n) $; a placeholder dictionary $ R = \left\{ (r_k : I_k) \right\}_{k = 1, \cdots, K} $, where $r_k$ is the predefined \emph{role} of that argument (e.g. \textsc{agent}, \textsc{patient}); and an index set $I_k \subseteq \{ 1, \cdots, n \}$, a set containing indices of tokens in the statement $S$ (i.e. if $I = \{i_1, \cdots, i_l\}$, then $(s_{i_1}, \cdots, s_{i_l})$ is a placeholder).\footnote{~Our bleached statements are inspired in part by linguistic resource creation efforts by \citet{white-18}.}
  An example bleached statement in the ACE 2005 dataset for the event type \textsc{Life:Die} (also used in our Figure~\ref{fig:statement-completion} for illustration purposes) is:
  \begin{eqnarray}
    \role[]{someone}{agent} ~{\rm killed}~ \role[]{someone~else}{victim} ~{\rm with}~ \role[]{something}{instrument} \nonumber \\
    ~{\rm in}~ \role[]{some place}{place} ~{\rm at}~ \role[]{some time}{time} \nonumber
  \end{eqnarray}
  This statement is accompanied by the following placeholder dictionary, in which each role is mapped to an index set that highlights the placeholder in the bleached statement\footnote{~The model itself does not see the role names. They are used only for human readability and evaluation.}:
  \begin{equation}
    R = \left[ \begin{array}{rcl}
      \textsc{agent}      &:& \{1\} \\
      \textsc{victim}     &:& \{3, 4\} \\
      \textsc{instrument} &:& \{6\} \\
      \textsc{place}      &:& \{8, 9\} \\
      \textsc{time}       &:& \{11, 12\} \\
    \end{array}\right] \nonumber
  \end{equation}
  
  The event extraction task as defined in the ACE 2005 dataset also requires finding an \emph{event trigger}---a span in the text that most clearly expresses the event's occurrence. In this example, the trigger is the word ``\emph{killed}''. For a consistent implementation, we consider the trigger to be a special argument of the event, with role name \textsc{trigger}.

  Formally, the task is: given a bleached statement $S$, its placeholder dictionary $R$, and text tokens $T$, return a dictionary $\hat{R}$ that contains the event trigger and the the extracted arguments. Such a result is shown in the bottom right of Figure~\ref{fig:statement-completion}. Note that the \textsc{instrument} role is not filled in the example because the model does not find a span to fill it.

\section{Approach}  

  Given a bleached statement with multiple placeholders, we do not fill the placeholders in parallel---instead, we fill them incrementally in an enforced order.\footnote{The enforced order is denoted in our examples by indices on the placeholders and the values that fill them.} In each step, the model attempts to fill a single focused placeholder, which is replaced by the extracted span(s) thereby creating a refined statement (see Figure~\ref{fig:statement-completion}).
  In this work we fill the placeholders in the statement \emph{from left to right} and leave other orders as future work.
  
  Formally, in each round, our model returns multiple arguments (see ``Multiple Argument Selector'' section below) for each placeholder:
  \begin{equation}
    A \gets \textsc{GetArgs}(S, I, T) \ , \nonumber
  \end{equation}
  where $S$ is the (partially refined) statement, $I$ is the index set that covers the focused placeholder (which corresponds to a role), and $T$ is the text to extract from. The returned argument set $A$ contains a number of text spans (potentially zero) in $T$ that replace the placeholder in $S$ picked out by $I$.
  
  \begin{algorithm}[t] 
    \caption{Argument extraction}
    \label{alg:arg-extraction}
    \begin{algorithmic}
      \Require {\small statement $S$, placeholder dictionary $R$, text $T$}
      \Ensure extracted argument structure $E$
      \Function{ExtractArgs}{$S, R, T$}
        \State $i \gets 1$  \Comment{$i$-th round}
        \State $S^{(1)} \gets S$ \Comment{the initial statement}
        \State $E \gets \varnothing$ \Comment{extracted event}
        \For{$(r, I) \in R$}
          \State $A \gets \textsc{GetArgs}(S^{(i)}, I, T)$
          \If{$A \ne \varnothing$}
            \State $S^{(i+1)} \gets$ replace the $I$ tokens in $S^{(i)}$ with $A$ \Comment{refine the statement}
            \State $E \gets E \cup ( r : A )$  
          \Else{ $S^{(i+1)} \gets S^{(i)}$} \Comment{skip to the next role}
          \EndIf
          \State $i \gets i + 1$
        \EndFor \\
        \Return $E$
      \EndFunction
    \end{algorithmic}
  \end{algorithm}
  
  If $A$ is the empty set, then the model did not find an appropriate text span to replace the placeholder. If the answer set $A$ is not empty, we replace the placeholder with the extracted span. Note that in some cases, there can be more than one argument that fits a role. Consider the following bleached statement (for the ACE 2005 event \textsc{Life:Marry}), focused on the first placeholder ``\emph{some people}'':
  \begin{equation}\resizebox{\columnwidth}{!}{$
      \role[1]{\bf some people}{person}~\textrm{married in}~\role[2]{some location}{place}~\textrm{at}~\role[3]{some time}{time} \nonumber$}
  \end{equation}
  
  We expect multiple arguments for the same role \textsc{person} in this event. If our model returns $A = \{\textrm{``Kim'', ``Pat''}\}$, i.e. a set containing multiple extracted arguments, we replace the placeholder with all arguments, concatenated with the ``\emph{and}'' token, and shift the focus to the next placeholder, creating the refined statement:
  \begin{equation}\resizebox{\columnwidth}{!}{$
    \role[1]{Kim and Pat}{person}~\textrm{married in}~\role[2]{\bf some location}{place}~\textrm{at}~\role[3]{some time}{time} \nonumber$}
  \end{equation}

  If our model returns nothing, i.e. $A = \varnothing$, we simply skip the placeholder and move the focus to the next placeholder. For example, if the model finds no argument for the \textsc{place} role, the refined statement of the next iteration would be
  \begin{equation}\resizebox{\columnwidth}{!}{$
    \role[1]{Kim and Pat}{person}~\textrm{married in}~\role[2]{some location}{place}~\textrm{at}~\role[3]{\bf some time}{time} \nonumber$}
  \end{equation}
  
  We run this iterative process until all roles of an event are visited. An advantage of this method is that during the incremental refinement process, the statement always remains a natural language sentence. The incremental process for extracting event arguments is formalized in Algorithm \ref{alg:arg-extraction}, given the initial bleached statement $S$, the role dictionary $R$, and the text $T$ to extract from.

  \begin{algorithm}[t]
    \caption{Event extraction}
    \label{alg:evt-extraction}
    \begin{algorithmic}
      \Require ontology $\mathcal{O}$, text $T$
      \Ensure extracted event structures $\mathcal{E}$
      \Function{ExtractEvents}{$\mathcal{O}, T$}
        \State $\mathcal{E} \gets \varnothing$
        \For {$(\tau, S, R) \in \mathcal{O}$}
          \State $\textit{triggers} \gets \textsc{TriggerId}(S, R, T)$
          \For {$t \in \textit{triggers}$}
            \State $S^\prime \gets \textsc{AnchorTrigger}(S, t)$
            \State $E \gets \textsc{ExtractArgs}(S^\prime, R, T)$
            \State $\hat{R} \gets E \cup (\textsc{trigger} : t)$
            \State $\mathcal{E} \gets \mathcal{E} \cup \hat{R}$
          \EndFor
        \EndFor \\
        \Return $\mathcal{E}$ \Comment{all events extracted from $T$}
      \EndFunction
    \end{algorithmic}
  \end{algorithm}
  
  Annotation manuals of interest usually define multiple event types. For each event type $\tau$ described in the manual, we require a bleached statement $S$ and a role dictionary $R$. These together form our ontology $\mathcal{O} = \{(\tau_k, S_k, R_k)\}$. To perform full event extraction (Algorithm \ref{alg:evt-extraction}), we first run a trigger detection model for \emph{all event types} (see ``Trigger Identification'' section below) specified in the ontology. For those event types whose trigger is found, we proceed with argument extraction (Algorithm \ref{alg:arg-extraction}).

  \subsection{Model}
  
    \paragraph{Architecture for MRC}
    In light of recent advancements in NLP from large-scale pre-training, we use BERT~\cite{devlin-etal-2019-bert} as our sequence encoder. We first review the answer selector architecture for machine reading comprehension (MRC) used in BERT, then extend it for our approach.
    
    Under the formulation of MRC, each training data point is of the form $(S, T)$ where $S$ is a natural language question with tokens $S = (s_1, \cdots, s_n)$ and $T$ is the text to extract answers from, with tokens $T = (t_1, \cdots, t_m)$. The model returns a span in $T$ or predicts that the question is not answerable, in which case an empty span is returned.
    
    To perform MRC, \citet{devlin-etal-2019-bert} proposed the following architecture. First the question $S$ and the text $T$ are concatenated with special delimiters and passed through the BERT contextualizer:
    \begin{equation}
        \mathrm{BERT}\left( [\textsc{cls}, s_1, \cdots, s_n, \textsc{sep}, t_1, \cdots, t_m, \textsc{sep}] \right) ,\nonumber
    \end{equation}
    where \textsc{cls} is a special sentinel token whose embedding encompasses the whole string, and \textsc{sep} is a sentence separator. We denote the output encoding of each question token $s_i ~(1 \le i \le n)$ as $\mathbf{s}_i \in \mathbb{R}^d$, and the encoding of each text token $t_j ~(1 \le j \le m)$ as $\mathbf{t}_j \in \mathbb{R}^d$.
    Additionally, two vectors, $\mathbf{b}_{\rm left}$ and $\mathbf{b}_{\rm right}$, for the left and right boundaries of the answer span are learned. The probability of each token $t_j ~(1 \le j \le m)$ being the left or right boundary of the answer span is computed as 
    \begin{equation}
        P_{\rm left}(t_j) \propto \exp(\mathbf{b}_{\rm left} \cdot \mathbf{t}_j); ~ P_{\rm right}(t_j) \propto \exp(\mathbf{b}_{\rm right} \cdot \mathbf{t}_j) \nonumber
    \end{equation}
    The two vectors $\mathbf{b}_{\rm left}$ and $\mathbf{b}_{\rm right}$ act as attention \emph{query vectors} to the text, resulting in a soft pointer over the text tokens.
    
    
    
    \paragraph{Multiple Argument Selector}
    Our scenario is fundamentally different from MRC in two ways: (1) Our query is not formulated as a natural language question; instead, it is a cloze-style problem with a natural language statement and a highlighted blank to fill; (2) For some cases, there can be more than one answer for a given blank. Previous MRC models support extracting only at most one answer.
    
    To accommodate these requirements, we propose a new architecture for this scenario that describes the \textsc{GetArgs} function in Algorithm \ref{alg:arg-extraction}. Given a bleached statement $S = (s_1, \cdots, s_n)$ with a highlighted placeholder span with indices $I = \{i_1, \cdots, i_l\} \subseteq \{ 1, \cdots, n \}$, instead of two attention query vectors $\mathbf{b}_{\rm left}$ and $\mathbf{b}_{\rm right}$ to get the left and right boundary for the answer span, we consider the problem of answer span selection as a \emph{tagging} problem, first proposed in \citet{yao-etal-2013-answer}, where answer spans are tagged using a linear chain CRF \cite{LaffertyMP01}. By considering answer span selection as tagging, our model selects potentially multiple spans for a query.
    
    We enforce the constraint that all extracted spans come from the same sentence in the text, but in general this constraint need not be enforced. Additionally, our model operates on single-sentence contexts, so information available in other sentences is not considered.
    
    We use the {\tt BIO} tagging scheme~\cite{ramshaw-marcus-1995-text}, where each token in the text is tagged with {\tt B} (beginning), {\tt I} (inside), or {\tt O} (outside). 
    In a linear-chain CRF, the probability of an output tag sequence $y_1, \cdots, y_j$ (for each $j$, $y_j \in \{ \texttt{B}, \texttt{I}, \texttt{O} \}$) given the text $T = (t_1, \cdots, t_j)$ is
    \begin{equation}
        \label{eq:crf}
        P(y_1,{\small\cdots},y_j | t_1, {\small\cdots}, t_j) \propto \prod_{j=1}^{m} \psi(y_{j-1}, y_j, j) \ ,
    \end{equation}
    where we define the potential function $\psi(y_{j-1}, y_j, j)$ as the output of a neural function described below. Our model is trained to maximize $P$.
    
    We first compute an attentive representation for a placeholder with respect to each text token $t_j$, using the attention mechanism proposed by \citet{luong-etal-2015-effective}, since the placeholder is of variable length but we desire a fixed-size vector representation:
    \begin{align}
        a_{ij} &= \frac{ \exp \left( \mathbf{s}_i \cdot \mathbf{t}_j \right) } { {\displaystyle\sum}_{i^\prime \in I}~{\exp \left( \mathbf{s}_{i^\prime} \cdot \mathbf{t}_j \right) }}  \\
        \tilde{\mathbf{s}}_j &= \sum_{i \in I} a_{ij} \mathbf{s}_i
    \end{align}
    Then the attentive placeholder representation $\tilde{\mathbf{s}}_j$, together with its corresponding text token representation $\mathbf{t}_j$, are joined using various matching methods proposed in \citet{mou-etal-2016-natural}:\footnote{$|\cdot|$ is elementwise absolute value, $\odot$ is elementwise product, and $[;]$ is vector concatenation.}
    \begin{equation}
        \mathbf{x}_j = \left[ \tilde{\mathbf{s}}_j ~;~ \mathbf{t}_j ~;~ |\tilde{\mathbf{s}}_j - \mathbf{t}_j| ~;~ \tilde{\mathbf{s}}_j \odot \mathbf{t}_j  \right]
    \end{equation}
    yielding the joined feature vector $\mathbf{x}_j \in \mathbb{R}^{4d}$.
    
    Finally the joined feature vector $\mathbf{x}_j$ is passed through a multi-layer feed-forward neural network to get the final potential function for each token and each predicted tag type $y_j \in \{ \texttt{B}, \texttt{I}, \texttt{O} \}$:
    \begin{equation}
        \psi(y_{j-1}, y_j, j) = \mathrm{FFNN}_{y_j} (\mathbf{x}_j)
    \end{equation}
    In our experiments, we pass $\mathbf{x}_j$ through 4 layers, with output dimensions $2d, d, d$, and $1$, respectively, and tanh as the nonlinearity function between layers.
 
    \paragraph{Trigger Identification}
    Triggers of events can be thought as a special argument, which usually is the main verb (or a nominalized verb) that expresses the occurrence of an event. We reuse the argument selection model for trigger identification: the highlighted token set for the trigger is all tokens in the statement that are not part of any standard argument:
    \begin{equation}
      I_{\textrm{trigger}} = \{1, \cdots, n\} \setminus \bigcup_{(r, I) \in R} I
    \end{equation}
    For example, the highlighted token set for the trigger of the statement in Figure~\ref{fig:statement-completion} consists of the tokens underlined below:
    \begin{quote}
      someone \underline{killed} someone else \underline{with} something \underline{in} some place \underline{at} some time
    \end{quote}

  \subsection{Training Data Generation}
    We generate data examples in the form of $(S, I, T)$ triples to train the argument extractor, where $S$ is a bleached statement, $I$ is the index set of the focused placeholder, and $T$ is the text. Algorithm \ref{alg:arg-extraction} generates a sequence of bleached statements, where each successive statement is a refinement of its predecessor. During training, instead of replacing placeholders with their predicted arguments $A \gets \textsc{GetArgs}(S, I, T)$, we replace them with the gold argument(s) from the event extraction dataset.
    
    \paragraph{Negative Sampling} For trigger identification, we augment each example with negative samples from the set of event types not found in the example's text. For each event, $\alpha$\% of the non-occurring event types are taken as negative samples. We tune $\alpha \in \{ 10, 20, 30, 40, 50 \}$.

  \subsection{Recasting MRC Data for Pre-training}
  
    \squad~\cite{rajpurkar-etal-2016-squad} is a reading comprehension dataset consisting of questions on a set of Wikipedia articles, where the answer to each question is a span of text extracted from the corresponding reading passage. Its version~2.0~\cite{rajpurkar-etal-2018-know} contains additional data that poses unanswerable questions to reading comprehension systems. To do well, a system should learn to abstain from answering when no answer is supported by the text.
    
    We employ recast versions of the training and development splits of \squadtwo~as pre-training data for our event extraction system. We cast each \squad~natural language question to a format similar to our bleached statements, where the \wh- question phrases of the questions are tagged as the placeholders to be filled. For example, given the following \squad~question,
    \begin{quote}
        \textit{\underline{What form of oxygen} is composed of 3 oxygen atoms?}
    \end{quote}
    the extracted \wh- phrase is ``\emph{What form of oxygen}'', which is chosen as the single placeholder in this statement. 
    This is answered as
    \begin{equation}
        \role[]{ozone}{answer}~{\textrm{is composed of 3 oxygen atoms?}} \nonumber
    \end{equation}
    
    This methodology is linguistically motivated, as both questions (as in \squad) and bleached statements (this work) reduce to logical forms with the same predicate. The denotations can be written (using a generic operator $Q$) as
    \begin{align*}
        Q x. ~& \llbracket \textrm{form of oxygen} \rrbracket(x)\\
                    &  \wedge  \llbracket \textrm{composed of three oxygen atoms} \rrbracket(x)
    \end{align*}
    where $Q$ is $\lambda$ for the question and is $\exists$ for the bleached statement. Hence the \wh- phrase is semantically similar to an existentially quantified phrase (e.g. \emph{some form of oxygen}, where \emph{some} introduces existential quantification), despite their pragmatic difference in illocutionary force (inquiring vs. stating). Additionally, \wh- phrases presuppose the existence of their answer referent; to use a \wh- phrase when no referent exists would be infelicitous. Hence \wh- question phrases serve the same function as the existentially quantified placeholder phrases in our bleached statements, and so the recast \squad~questions are appropriate data for pre-training.
    
    We extract \wh- phrases through syntactic analysis of the questions. We define the \wh- phrase of a question to be the maximum span in its constituency parse that bears any of the question tags in the Penn Treebank~\cite{MarcusSM94} parsing annotation guideline.\footnote{These include  the following tags (followed by examples): \texttt{WHADJP} (how many), \texttt{WHADVP} (why), \texttt{WHNP} (which book), texttt{WHPP} (by whose authority), \texttt{WDT} (which), \texttt{WP} (who), \texttt{WP\$} (whose), \texttt{WRB} (where).}
    
    We employ the neural span-based constituency parser~\cite{stern-etal-2017-minimal} in the AllenNLP~\cite{AllenNLP} toolkit to parse the \squad~questions for extracting the \wh- phrases.

\section{Experiments and Discussions}

  \subsection{Event Extraction on ACE 2005}
    
      \begin{table*}[t!]
        \centering\adjustbox{max width=\linewidth}{
        \begin{tabular}{rcccccccccccc}
          \toprule
          {\bf Model} & \multicolumn{6}{c}{\bf Trigger} & \multicolumn{6}{c}{\bf Argument} \\
          \cmidrule(lr){2-7} \cmidrule(lr){8-13}
          & \multicolumn{3}{c}{\bf Identification} & \multicolumn{3}{c}{\bf Classification} & \multicolumn{3}{c}{\bf Identification} & \multicolumn{3}{c}{\bf Classification} \\
          \cmidrule(lr){2-4} \cmidrule(lr){5-7} \cmidrule(lr){8-10} \cmidrule(lr){11-13}
          & P & R & $\rm F_1$ & P & R & $\rm F_1$ & P & R & $\rm F_1$ & P & R & $\rm F_1$ \\
          \midrule
          \textsc{JointFeature} & \best{77.6} & 65.4 & 71.0 & \best{75.1} & 63.3 & 68.7 & \best{73.7} & 38.5 & 50.6 & \best{70.6} & 36.9 & 48.4 \\
          \textsc{Joint3EE} & 70.5 & 74.5 & 72.5 & 68.0 & 71.8 & 69.8 & 59.9 & \best{59.8} & \best{59.9} & 52.1 & \best{52.1} & \best{52.1} \\
          \midrule
          Ours w/ partial data & 64.5 & 62.3 & 63.4 & 60.9 & 58.7 & 59.8 & 43.1 & 42.8 & 43.0 & 35.2 & 34.9 & 35.0 \\
          Ours w/o pre-training & 50.0 & \best{85.7} & 63.2 & 48.1 & \best{82.4} & 60.7 & 29.3 & 55.0 & 38.2 & 24.7 & 46.4 & 32.2 \\
          \midrule
          Ours w/ full data & 68.9 & 77.3 & \best{72.9} & 66.7 & 74.7 & \best{70.5} & 44.9 & 41.2 & 43.0 & 44.3 & 40.7 & 42.4 \\
          \bottomrule
        \end{tabular}}
        \caption{P(recision), R(ecall), and $\rm F_1$ obtained by models on the ACE 2005 dataset. Best results are \best{bolded}. Using the full training set improves $\rm F_1$ performance over using the partial training set on all metrics except argument identification (equal). Pre-training on the recast \squadtwo~dataset improves $\rm F_1$ performance over no pre-training on all metrics.}
        \label{tab:aceresults}
      \end{table*}
    We evaluate our approach on the ACE 2005 dataset and use the same data splits as previous work, in which 40 newswire documents are used as the test set, another 30 documents of different genres are selected as the development set, and the remaining 529 documents constitute the training set~\cite{li-etal-2013-joint,yang-mitchell-2016-joint,NguyenN19}. Following previous work, we use four evaluation metrics: (1) \textit{Trigger Identification}: a trigger is correctly identified if its span offsets exactly match a reference trigger; (2) \textit{Trigger Classification}: a trigger is correctly classified if its span offsets and event subtype exactly match a reference trigger; (3) \textit{Argument Identification}: an argument is correctly identified if its span offsets and corresponding event subtype exactly match a reference argument; and (4) \textit{Argument Classification}: an argument is correctly classified if its span offsets, corresponding event subtype, and argument role exactly match a reference argument.
    The overall performance is evaluated using \textit{precision} (P), \textit{recall} (R), and \textit{F-measure} (${\rm F}_1$) for each metric.
    
    
    We use BERT for sequence encoding.\footnote{We use the ~\textsc{bert-base-cased} model, which has $12$ layers, $768$-dimensional hidden embeddings, $12$ attention heads, and $110$ million parameters.} For pre-training on the recast \squadtwo~dataset, we follow the previously mentioned pre-processing strategy. We pre-train on the training set of \squadtwo~and perform early stopping using the development partition. Examples that do not have exactly 1 \wh-phrase are discarded.\footnote{Our effective training set contains 128,649 examples after 1,670 were discarded, and the effective development set contains 11,772 examples after 100 were discarded.} The maximum sequence length is 512 word pieces, the maximum query length is 128 word pieces, the learning rate is $3\times 10^{-5}$ with an Adam optimizer, the maximum gradient norm for gradient clipping is set to $1.0$, and the number of training epochs is $3$.
    
    
    
    After pre-training on \squadtwo, we fine-tune the model on ACE 2005. While keeping other hyperparameters unchanged, we set the learning rate to $1\times 10^{-5}$ and the number of training epochs to $8$. During fine-tuning, we employ negative sampling and set the negative sampling rate to $30\%$.
    
    In addition to fine-tuning on the full training set of ACE 2005, we consider a single-genre ``partial'' training setting in which the model is trained only on the 58 documents that appear in the newswire portion of the full training set.
    
    \paragraph{Experimental Results \& Discussion}

      We train our model using full and partial training data and compare with two joint event extraction model baselines. The \textsc{JointFeature} model~\cite{yang-mitchell-2016-joint} is a feature-based model that exploits document-level information; the \textsc{Joint3EE} model~\cite{NguyenN19} is a neural model that achieves state-of-the-art performance on ACE 2005. These two models represent the state-of-the-art performance for feature-based and neural models, respectively.
      
      Table~\ref{tab:aceresults} reports the performance of the systems on the four evaluation metrics. Training on the full training set improves $\rm F_1$ performance over training on the partial training set, giving the largest improvement on trigger identification and classification. Additionally, pre-training on the recast \squadtwo~dataset provides large $\rm F_1$ improvements (4.8\%--10.2\% absolute $\rm F_1$ increase) on all four evaluation metrics. Our model also tends to have higher recall than precision, especially on trigger identification and classification, and suffers from low precision compared to prior work. Our model achieves state-of-the-art performance on trigger identification and trigger classification.
    
      Because our model does not explicitly incorporate entity mention detection, we hypothesize that our MRC-inspired approach predicts answer spans that are semantically correct but do not exactly match the gold answers, hurting performance on argument-related subtasks. We compare predicted arguments with gold references and find the following sources of errors:
      
      \begin{itemize}
          \item \textit{Relative clauses}: Our model predicts \textit{Mosul} whereas the gold answer is \textit{Mosul, where U.S. troops killed 17 people in clashes earlier in the week}.
          \item \textit{Counts}: The gold annotation is \textit{300 billion yen} but our model predicts \textit{300 billion}.
          \item \textit{Durations}: The gold annotation is \textit{lasted two hours} but our model predicts \textit{two hours}.
          
      \end{itemize}
    

  \subsection{Few-shot Learning on FrameNet}
    
    In order to evaluate our approach under a lower-resource setting than the partial training setting, we consider few- and zero-shot learning on annotated documents from FrameNet.\footnote{We use the Full Text Annotation portion of FrameNet.} $7$ documents are excluded from FrameNet's $107$ annotated documents because they do not have frame annotations. The arguments in our bleached statements consist of only the frame's core \emph{frame elements}. We use the same four evaluation metrics as used for ACE 2005.
    
    We split the remaining $100$ documents into $5$ training documents and $95$ test documents and experiment with training on between $1$ and $5$ documents.\footnote{Because we do not perform a hyperparameter sweep in this setting, we do not use a development set.} We pick the top-$10$ most frequent frames that represent events and write bleached statements based on their frame definitions. 
    
    Following the same pre-training setup, we then train the model using the same hyperparameters as the model fine-tuned on ACE 2005.
    
    \paragraph{Experimental Results \& Discussion}
    
    
    \begin{figure}[t]
      \centering
      \includegraphics[width=0.9\linewidth]{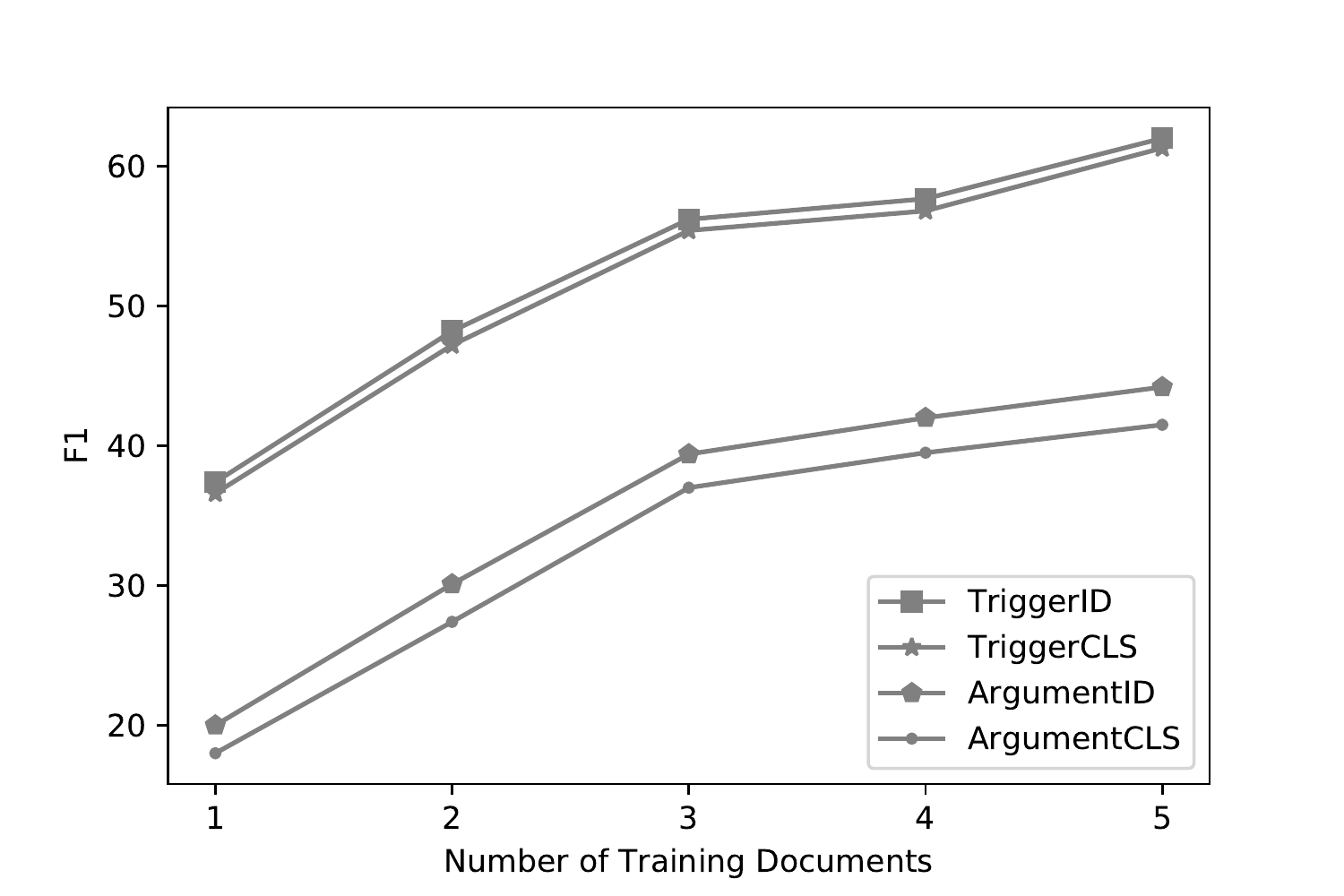}
      \caption{Few-shot experimental results on FrameNet.}
    \label{fig:fn-few-shot}
    \end{figure}

    The results in Figure~\ref{fig:fn-few-shot} show that $\rm F_1$ performance on all evaluation metrics increases as more documents are added to the training set. The marginal utility of adding training documents almost monotonically decreases as the training set increases, so that performance from training on 3 documents roughly matches performance from training on 5 documents.
    
  \subsection{Zero-shot Learning on FrameNet}
    
    We additionally investigate the model's ability to generalize to unseen event types using the same dataset as the few-shot setting. We employ a leave-one-out strategy to the frames in Table~\ref{tab:fn-zero-shot}, training on 9 frames and testing on the other 1.
    
    \paragraph{Experimental Results \& Discussion}
    
    \begin{table}[t!]
      \centering\adjustbox{max width=\linewidth}{
        \begin{tabular}{rrrrr}
          \toprule
          {\bf Frame} & \multicolumn{2}{c}{\bf Trigger} & \multicolumn{2}{c}{\bf Argument} \\
          \cmidrule(lr){2-3} \cmidrule(lr){4-5}
          & \bf Id. & \bf Cls. & \bf Id. & \bf Cls. \\
          \midrule
             \sc Arriving & $9.1$ & $9.1$ & $6.1$ & $6.1$ \\
             \sc Attack & $66.4$ & $66.4$ & $42.1$ & $42.1$ \\
             \sc Getting & $17.6$ & $17.6$ & $14.3$ & $13.8$ \\
             \sc Intentionally\_create & $12.5$ & $12.5$ & $7.1$ & $7.1$ \\
             \sc Killing & $22.4$ & $22.4$ & $13.1$ & $7.0$ \\
             \sc Manufacturing & $37.4$ & $37.4$ & $27.6$ & $26.7$ \\
             \sc Scrutiny & $16.6$ & $16.6$ & $11.7$ & $11.0$ \\
             \sc Statement & $0.6$ & $0.6$ & $0.7$ & $0.5$ \\
             \sc Supply & $3.1$ & $3.1$ & $1.7$ & $1.5$ \\
             \sc Using & $10.4$ & $10.4$ & $10.0$ & $8.8$ \\
          \midrule
          Macro-averaged $\rm F_1$ & $19.6$ & $19.6$ & $13.4$ & $12.46$ \\
          \bottomrule
        \end{tabular}}
      \caption{Zero-shot results on FrameNet frames. Trigger classification performance is equivalent to trigger identification performance because we evaluate on only one frame in each zero-shot learning experiment.}
      \label{tab:fn-zero-shot}
    \end{table}

    The results in Table~\ref{tab:fn-zero-shot} reveal a large variation in performance on the frames. The best performance is achieved on the \textsc{Attack} frame, but frames such as \textsc{Statement} achieve poor performance. 
    
    We report the \textit{macro-averaged} $\rm F_1$ over all frames to reveal overall performance instead of \textit{micro-average}, since we care how the approach generalizes to different frames. Overall, the macro-averaged $\rm F_1$ shows that the model can feasibly extract information about events of unseen types, but performance varies greatly across frames.
    
    Possible reasons why the \textsc{Statement} frame has low performance include: (1) the event type being too general, (2) the bleached statement being poorly constructed, (3) the span for the ``message'' role being long and difficult to tag exactly correctly using the {\tt BIO} scheme.


\section{Conclusion \& Future Work}
We present an approach to event extraction that uses bleached statements to give a model access to information contained in annotation manuals. Our model incrementally refines the statements with values extracted from text. We also demonstrate the feasibility of making predictions on event types seen rarely or not at all. Future work can apply our approach to $n$-ary relation extraction.

\section*{Acknowledgments}

This research was supported by the JHU HLTCOE, DARPA AIDA, DARPA KAIROS, IARPA BETTER, and NSF-BCS (1748969/1749025).  The U.S. Government is authorized to reproduce and distribute reprints for Governmental purposes. The views and conclusions contained in this publication are those of the authors and should not be interpreted as representing official policies or endorsements of DARPA or the U.S. Government.

\bibliography{aacl-ijcnlp2020}
\bibliographystyle{acl_natbib}

\appendix

\end{document}